\documentclass{article}

\usepackage{PRIMEarxiv}

\usepackage[utf8]{inputenc} 
\usepackage[T1]{fontenc}    
\usepackage{hyperref}       
\usepackage{url}            
\usepackage{booktabs}       
\usepackage{amsfonts}       
\usepackage{nicefrac}       
\usepackage{microtype}      
\usepackage{lipsum}
\usepackage{fancyhdr}       
\usepackage{graphicx}       

\usepackage{subcaption}
\usepackage{amsmath}
\usepackage{algorithm}
\usepackage{multirow}
\usepackage{algorithmic}
\graphicspath{{media/}}     
\usepackage{natbib}
\setcitestyle{authoryear,round,citesep={;},aysep={,},yysep={;}}

\title{Temporally Layered Architecture for Adaptive, Distributed and Continuous Control
}

\author{
  Devdhar Patel \\
University of Massachusetts Amherst \\
   Amherst, MA 01003, USA\\
  \texttt{devdharpatel@cs.umass.edu} \\
   \And
  Joshua Russell \\
    College of Computer and Information Sciences, \\
University of Massachusetts Amherst \\
   Amherst, MA 01003, USA\\
  \texttt{jgrussell@cs.umass.edu } \\
  \AND
  Francesca Walsh \\
  University of Massachusetts Amherst \\
   Amherst, MA 01003, USA\\
  \texttt{fnwalsh@umass.edu} \\
  \And
  Tauhidur Rahman \\
  University of California San Diego \\
  La Jolla, CA 92093, USA \\
  \texttt{trahman@ucsd.edu} \\
  \And
  Terrence Sejnowski \\
  	Salk Institute for Biological Studies \\
  La Jolla, CA 92037, United States \\
  \texttt{terry@salk.edu} \\
  \And
  Hava Siegelmann \\
University of Massachusetts Amherst \\
   Amherst, MA 01003, USA\\
  \texttt{hava@cs.umass.edu} \\
}

\begin{document}

\maketitle

\begin{abstract}
We present temporally layered architecture (TLA), a biologically inspired system for temporally adaptive distributed control. TLA layers a fast and a slow controller together to achieve temporal abstraction that allows each layer to focus on a different time-scale. Our design is biologically inspired and draws on the architecture of the human brain which executes actions at different timescales depending on the environment’s demands. Such distributed control design is widespread across biological systems because it increases survivability and accuracy in certain and uncertain environments. We demonstrate that TLA can provide many advantages over existing approaches, including persistent exploration, adaptive control, explainable temporal behavior, compute efficiency and distributed control. We present two different algorithms for training TLA: (a) Closed-loop control, where the fast controller is trained over a pre-trained slow controller, allowing better exploration for the fast controller and closed-loop control where the fast controller decides whether to "act-or-not" at each timestep; and (b) Partially open loop control, where the slow controller is trained over a pre-trained fast controller, allowing for open loop-control where the slow controller picks a temporally extended action or defers the next n-actions to the fast controller. We evaluated our method on a suite of continuous control tasks and demonstrate the advantages of TLA over several strong baselines.
\end{abstract}

\keywords{Reinforcement Learning \and Adaptive Control \and Real-Time \and Continuous Control}

\section{Introduction}

\begin{figure*}
\begin{subfigure}{0.45\textwidth}
    \includegraphics[width=\textwidth]{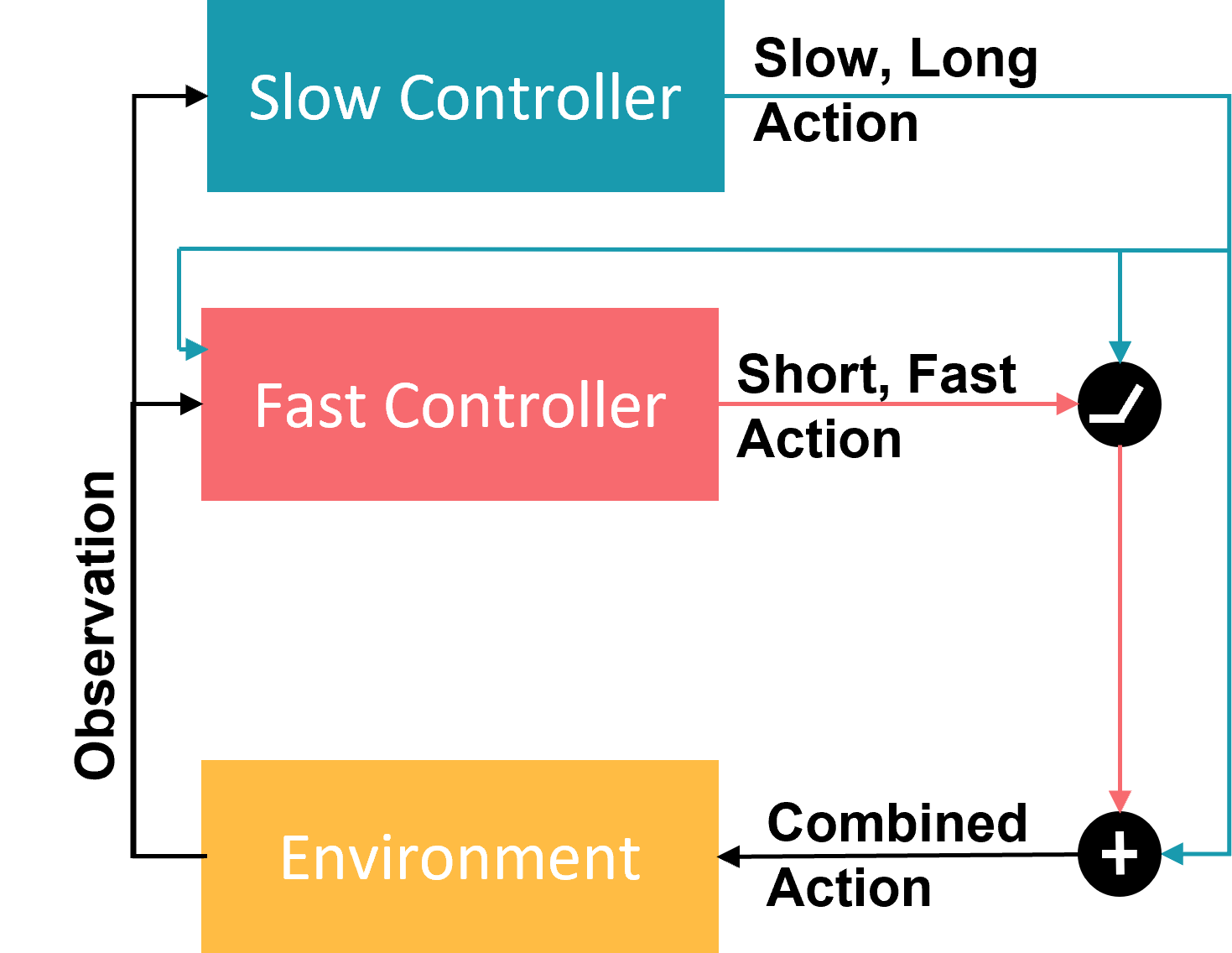}
    \caption{Closed Loop Control}
    \label{fig:clc}
\end{subfigure}
\begin{subfigure}{0.45\textwidth}
    \includegraphics[width=\textwidth]{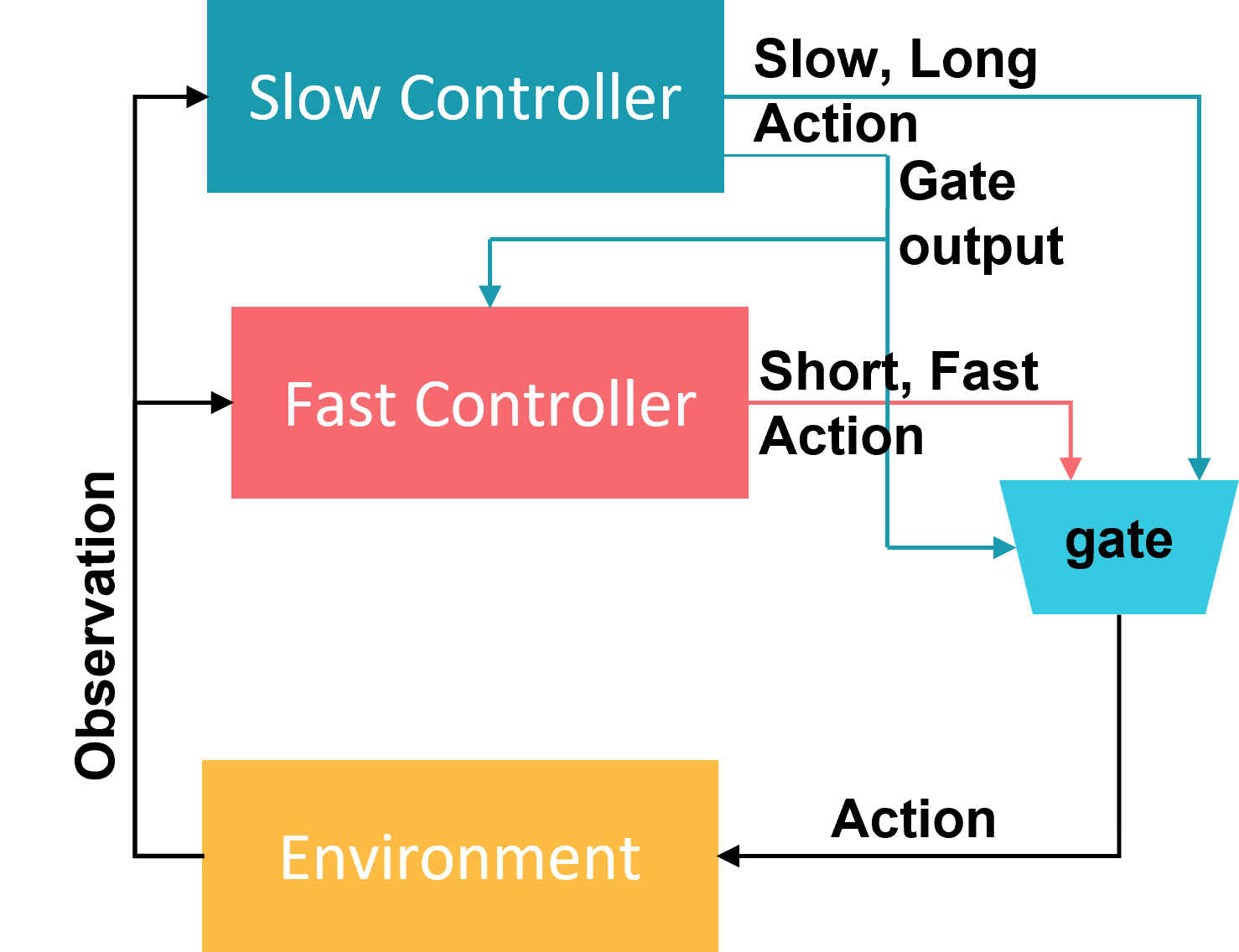}
    \caption{Open Loop Control}
    \label{fig:olc}
\end{subfigure}
\caption{Two variants of temporally layered architecture: (a) Closed Loop Control: where the fast controller acts on top of the slow controller at each time step. This allows the fast actions to be threshold gated for repetitions of the slow, long actions for multiple time-steps. (b) Open Loop Control: where the slow controller is trained on top of the fast controller allowing it to take longer actions by blocking the fast controller from processing inputs.}
\label{fig1}
\end{figure*}
Deep reinforcement learning (RL) has seen remarkable success on control tasks, even beating humans in real time games. However, most of these successes come from an RL agent acting at a constant frequency. Usually, this frequency is much faster than the average human response time \cite{OpenAI_dota}. However, a higher response frequency increases the energy consumption for processing inputs and performing actions. Fast actions also cause a more jerky behaviour, which might cause damage or discomfort to humans interacting with the RL agent. Additionally, a faster response time increases the task horizon, making it more difficult to train the agent.

On the other hand, a slower response time forces the RL agent to take longer ``macro'' actions without supervision. In continuous-time environments, this might lead to poor performance since the agent is now forced to plan ahead instead of being able to react to changes in the environment. Optimizing control also becomes more challenging if the macro actions are sequences of potentially 
different actions, as the number of macro actions becomes exponential in the length of the action sequence.

Thus, in RL algorithms, response time (timestep, frame-skip, action-repetition) is a very important hyper-parameter whose optimal value might vary significantly between different environments \cite{Braylan2015FrameSI}. In complex environments, the optimal value might also vary within an episode and vary significantly across episodes. In this paper, we propose a framework to leverage this phenomena by allowing the RL agent to adapt its action-frequency.

We take inspiration from the brain which has evolved to respond to many types of environments and accurately react in many different situations. The brain's design enables it to use context to modulate its response time and to accurately respond to a number of familiar and unfamiliar complex environments. This design allows it to conserve energy in situations where reacting slower is acceptable, while also acting faster when necessary. Building on the history of researching the speed/accuracy trade-off \cite{heitz2014speed}, recent work has shown that the brain might use distributed control to allow multiple independent systems to process the environment and accurately react. This distributed control allows for multiple layers of the biological neural network to activate and control muscle groups to execute complex behaviors. This enables the brain and central nervous system to trade-off between speed and accuracy as the situation demands \cite{Nakahira2021DiversityenabledSS}.


Inspired by this biological design, we propose Temporally Layered Architecture (TLA) (Fig. \ref{fig1}): a reinforcement learning architecture that layers two different networks with different response frequencies to achieve distributed temporally adaptive behavior. To avoid an exponential increase in the number of actions, each network has a constant response frequency - one fast and one slow. However, the RL agent can use their combination to adapt its response frequency. The temporally layered architecture allows TLA to easily abstract hierarchical temporal knowledge into layers that focus on different time-frames. Additionally, the distributed nature of the architecture makes it viable for distributed settings like FOG computing. Our contributions are:
\begin{enumerate}
    \item We propose a bio-inspired Temporally Layered Architecture (TLA), an alternative to classical RL algorithms that allows each layer to focus on a different temporal context.
    \item We introduce two algorithms for training TLA. Each providing a different advantage over traditional deep RL approaches.
    \item We experiment on continuous control tasks and demonstrate faster convergence, improved performance, fewer decisions and higher action repetition on all environments tested. In the partially open loop setting, we also demonstrate comparable performance to previous work. 
    \item We also show robust performance in a simulation delayed setting where each layer is operating at a different delay, demonstrating the viability of our approach for a distributed real-time setting with communication or other delays.
\end{enumerate}

\section{Background}
\subsection{Reinforcement learning}
The goal of reinforcement learning is to learn a reward maximizing behavior for the environment. In a standard reinforcement learning setting, the environment is represented as a Markov Decision Process (MDP). The MDP is characterized by a 6-tuple $(\mathcal{S}, \mathcal{A}, p, R, d_0, \gamma)$, where: $\mathcal{S}$ is a set of states, $\mathcal{A}$ is a set of actions, $p: \mathcal{S} \times \mathcal{A} \times \mathcal{S} \rightarrow [0,1]$ is a transition function, $R: \mathcal{S} \times \mathcal{A} \rightarrow \mathbb{R}$ is a reward function, $d_0: \mathcal{S} \rightarrow [0,1]$ is an initial state distribution, and $\gamma \in [0,1]$ is the discount factor.

The agent is characterized by a policy $\pi:  \mathcal{S} \times \mathcal{A} \rightarrow [0,1]$.
\begin{equation}
    \pi(s,a) := \Pr(A_t=a | S_t=s)
\end{equation}

The objective of the agent is to learn the optimal policy $\pi^*_{\phi}$, with parameters $\phi$, that maximises the expected sum of discounted rewards:
\begin{equation}
    J(\pi) := \mathbb{E}\left[\sum_{t=0}^{\infty}\gamma^t R_t | \pi\right] := \mathbb{E}\left[G_t | \pi\right]
\end{equation}

where $R_t$ is the reward at time $t$, and $G_t$ is the return from time $t$. The policy is updated using the gradient of the expected return $\nabla_{\phi}J(\pi_\phi)$. In actor-critic methods, the actor (policy) is updated using the policy gradient algorithm \cite{Silver2014DeterministicPG}:

\begin{equation}
    \nabla_\phi J(\pi_\phi) = \mathbb{E}[\nabla_\phi \pi_\phi (s) \nabla_a Q^\pi(s,a)|_{a=\pi_\phi{s}}]
\end{equation}

where $Q^\pi(s,a):= \mathbb{E}(G_t|S_t=s, A_t = a, \pi)$ is the action-value function or the Q-function. The Q-function gives the expected return of performing action $a$ at state $s$ and following the policy $\pi$ thereafter. This value is learned by the critic using temporal difference learning \cite{sutton2018reinforcement}:

\begin{equation}
    Q(S_t, A_t) \leftarrow Q(S_t, A_t) + \alpha \big[ R_{t} + \gamma \max_a Q(S_{t+1}, a) - Q(S_t, A_t)\big]
\end{equation}

where $\alpha \in [0, 1]$ is the learning rate. In deep Q-learning, the value is approximated using a differentiable function approximator $Q_\theta(s,a)$. In this paper, we use the twin-delayed deterministic policy gradient (TD3) method \cite{Fujimoto2018AddressingFA} that learns two Q-functions (critics) and uses the pessimistic value of the two for training a policy that is updated less frequently than the critics.

\subsection{Delayed Reinforcement Learning}

Unlike the traditional turn-based setting where the environment moves forward after the action is performed in the environment, in a real-time environment, the environment moves forward in time while the agent is picks an action. Thus, from the perspective of the agent, there is always a delay between the time when the action is picked and when the action is performed in the environment. This means that the agent needs to be aware of this delay or have compensatory policies, such as spinal cord reflexes observed in biological agents \cite{sandrini2005lower}. These policies and temporally layered responses allow the agent to adapt and survive in a real-time environment where such sensory processing delays can easily cause detrimental damage to the agent.

If the agent is continuously processing states and picking actions, its action frequency will be equal to its processing time. Thus, the delay is modeled to be equal to one time step, $t$, which is the time required to pick an action given an input state. 

In a delayed setting, the action, $a_{t+1}$, which is picked by processing the state, $s_t$, is then performed at state $s_{t+1}$. We use the Real-Time Markov Reward Process (RTMRP)  formulation proposed in \cite{Ramstedt2019RealTimeRL} and augment the state space by appending the action: $\mathcal{X} = \mathcal{S} \times \mathcal{A}$. While the environment transition function remains the same, at time $t$, the agent is unable to view the current state, therefore the delayed policy becomes:
\begin{equation}
    \bar{\pi} (a' | s, a):= \Pr(A_t = a' | S_{t-1} = s,  A_{t-1} = a)
\end{equation}

Thus, a fast actor with a lower $t$ will have a smaller delay between sensing and actuation. Moreover, action $t$ will persist for a shorter period of time.

\subsection{Timestep}

In most reinforcement learning (RL) problems, the agent's timestep is constant and often defined by the environment. In simulation, timestep is related to the response time of the agent, as a longer timestep would mean that the agent would be unable to respond to fast changes in the environment. We use timestep and response time interchangeably in this section.  Even when considering an agent with a constant response time, many other aspects of the control problem vary with the choice of its value:
\begin{enumerate}
    \item \textbf{Performance:} Generally, as the length of a timestep is decreased, the agent's performance on the reinforcement learning task is non-decreasing. This is intuitive since a faster agent can quickly react to environmental changes. However, a faster response speed also means that the agent slices an episode into more states resulting in a longer task horizon. This can decrease action-value propagation and can in turn slow convergence to optimal performance \cite{McGovern1997RolesOM}.
    \item \textbf{Energy:} Faster response time means processing more inputs per unit time and faster actuation, both of which require more energy expenditure. In an energy-constrained setting, the energy will bound the response speed of the agent. 
    \item \textbf{Memory Size:} Deep reinforcement learning algorithms rely on experience replay memory during learning \cite{Mnih2015HumanlevelCT}. A faster response time would result in the creation of more memories per unit time. Thus, a small memory size would bottleneck the performance and might even result in lower performance when the response speed increases. On the flip side, when the memory size is constrained, a slower response time might result in more efficient memory use. 
    \item \textbf{Network Size/Network Complexity:} Assuming that the agent uses a neural network to learn the policy, the neural network size would control the learned policy's complexity. A small neural network would result in a faster processing time but would only be able to learn a simple policy. In contrast, a larger neural network would increase the processing time but will be able to learn more complex policies. However, as described in the following paragraph, policy complexity increases when the processing time (and thus the response time) is decreased.
    \item \textbf{Reward Distribution}: Generally, reward is considered to be a property of the environment. Reward functions are often modeled so that reward is gained for reaching a certain state of the environment (important states are often referred to as the goal state or failure state). In such a setting, the return for an episode is independent of the response time of the agent. However, from the perspective of the agent, the temporal density of the reward (reward/state transitions per unit time) would decrease with a decrease in response time (as the agent becomes faster) as the total state transitions increase. This results in the task horizon increasing, making the RL problem more difficult. Especially for environments with sparse rewards where only the goal state has a positive reward, and all other states give a zero reward, a faster agent would have to explore more zero-reward state-action transitions before finally reaching the goal state (e.g., the mountain-car problem \cite{Moore90efficientmemory-based}). 

\section{Related Work}

\end{enumerate}
The idea of combining multiple controllers with different response times is, to the best of our knowledge, novel. However, it is related to several sub-fields of AI reviewed below: 

\subsection{Action repetition and frame skipping}
Frame-skipping and action repetition has been used as a form of partial open-loop control where the agent selects a sequence of actions to be executed without heeding the intermediate states. \citet{Hansen1996ReinforcementLF} proposed a mixed-loop control setting where sensing incurs a cost, thus allowing the agent to perform a sequence of actions to reduce the sensing cost. However, reinforcement learning with a sequence of actions is challenging since the number of possible action sequences of length $l$ is exponential in $l$. As a result, research in the area focuses on pruning the possible number of actions and states \cite{Hansen1996ReinforcementLF, Tan1991CostSensitiveRL, McCallum1996ReinforcementLW}. To avoid the exponential number of action sequences, other works have restricted the action sequences to be a sequence of the same action. The number of possible actions are therefore linear in $l$ \cite{Buckland1993TransitionPD, Kalyanakrishnan2021AnAO, Srinivas2017DynamicAR, Biedenkapp2021TempoRLLW, Sharma2017LearningTR}. \citet{Biedenkapp2021TempoRLLW} demonstrated that learning an additional action-repetition policy,which decides on the number of time steps to repeat a chosen action, can lead to faster learning and
reduce the number of action decision points during an episode. We use their approach as a benchmark in our experiments section.

\citet{McGovern1997RolesOM} analysed the effects of macro-actions and identified two advantages: improved exploration and faster learning due to reduced task horizon. \citet{Randlv1998LearningMI} also demonstrated empirically that macro actions significantly reduce the training time. \citet{Braylan2015FrameSI} showed that the performance of the DQN algorithm \cite{Mnih2015HumanlevelCT} on some Atari games can be significantly improved if the amount of frames skipped is increased.

However, these approaches require an environment that is predictable so that a sequence of actions can be planned and safely performed in the environment without further supervision. Additionally, as shown in this work, when applied to continuous domain, these approaches often require additional exploration. In contrast, the layered architecture in our approach allows the faster layer to monitor and act when required while the slower layer can be viewed as performing macro-actions.

Recently, \cite{Yu2021TAACTA} demonstrated a closed-loop temporal abstraction method on continuous domain using a "act-or-repeat" decision after the action is picked. However, their approach makes use of the state-action value from the critic thus requiring two forward passes of the critic in addition to the actor and the decision networks.

\subsection{Residual and Layered RL}

Recently, \citet{Jacq2022LazyMDPsTI} proposed Lazy-MDPs where the RL agent is trained on top of a sub-optimal base policy to act only when needed while deferring the rest of the actions to the base policy. They demonstrated that this approach makes the RL agent more interpretable as the states in which the agent chooses to act are deemed important. Similarly, for continuous environments, residual RL approaches learn a residual policy over a sub-optimal base policy so that the final action is the addition of both actions \cite{Silver2018ResidualPL, Johannink2019ResidualRL}. Residual RL approaches have demonstrated better performance and faster training. Our approach can be interpreted as temporal residual learning where a faster frequency network is trained on top of a (trained) slower frequency network to gain the benefits of macro-actions and residual learning.
    
   

\subsection{Delayed-aware and real-time RL}

The field of delayed MDPs has been well studied over the years \cite{Brooks1972TechnicalN}. \citet{Katsikopoulos2003MarkovDP} showed that observation delays and action delays are the same from the perspective of the agent and a MDP with delay can be reduced to a MDP without delays by augmenting the state space. \citet{Travnik2018ReactiveRL} proposed an algorithm to reduce the reaction time to improve its safety in a real-time environment. \citet{Firoiu2018AtHS} demonstrated that deep RL fails when it is forced to play Atari games at human reaction time. They propose that the agent learn a predictive model of the environment to undo the delay. 

\citet{Ramstedt2019RealTimeRL} used the state-augmented MDPs to create delay-aware versions of soft-actor-critic algorithm \cite{Haarnoja2018SoftAA} and demonstrated performance on real-time control tasks. \cite{Chen2021DelayAwareMR} used a similar setup to demonstrate delay-aware model-based RL performance on real-time control tasks. In our work, we use the same setting for the environments, however, since we have multiple controllers, the states are augmented by their combined actions.

\subsection{Options Framework}

The options framework \cite{Precup2000TemporalAI} is a common framework for temporal abstraction in RL. Options are defined as 3-tuples $\langle \mathcal{I}, \pi, \beta \rangle$. Where $\mathcal{I}$ is the set of initiation states which defines in which states the option can start; $\pi$ is the option policy that is followed for the duration of the option; and $\beta$ defines the probability of option termination in any given state. Options require prior knowledge about the environment to be defined. However, recent work have demonstrated that options that are automatically discovered by using the successor representation \cite{Machado2021TemporalAI} or the connectedness graph \cite{Chaganty2012LearningIA} can help improve exploration and thus improve learning. In a similar vein of research, \citet{Dabney2021TemporallyExtendedE} demonstrated that temporally extended actions improve exploration. Our work takes advantage of this phenomenon by layering a fast network taking short actions over a slow network that takes temporally extended actions.

\section{Methods}

\begin{figure*}[t]
\begin{subfigure}{0.3\textwidth}
    \includegraphics[width=\textwidth]{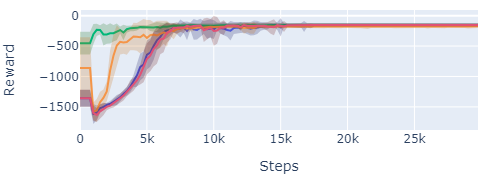}
    \caption{Pendulum-v1}
    \label{fig:lc1}
\end{subfigure}
\hfill
\begin{subfigure}{0.3\textwidth}
    \includegraphics[width=\textwidth]{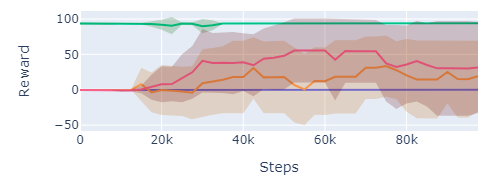}
    \caption{MountainCarContinuous-v0}
    \label{fig:lc2}
\end{subfigure}
\hfill
\begin{subfigure}{0.37\textwidth}
    \includegraphics[width=\textwidth]{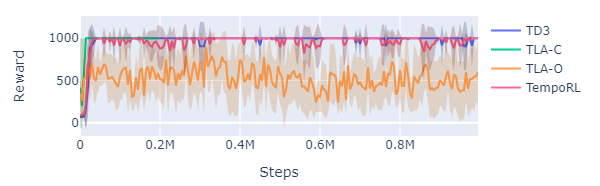}
    \caption{InvertedPendulum-v2}
    \label{fig:lc3}
\end{subfigure}
\hfill
\caption{Learning curves. The shaded region shows a single standard deviation over 5 trials.}
\label{fig:lc}
\end{figure*}

In this section, we describe two algorithms for training the temporally layered architecture (TLA), as shown in Fig. \ref{fig1}. The TLA  consists of two deep RL networks acting at different timesteps in order to abstract different temporal information about the task. In our experiments, for simplicity, the slower timestep is picked to be a multiple of the faster timestep: $t_{slow} = n \cdot t_{fast}$. Thus, for each action the slow agent picks, the fast agent picks $n$ actions. 

\subsection{Closed-loop temporally layered architecture}
In this setting, the slow network is pre-trained at a larger timestep. A larger timestep allows the network to better explore the environment and thus converge to an optimal policy faster. However, the optimal policy for the slow network might achieve a lower average return due to the longer timesteps. We then train the fast network on top of it so that the faster agent is acting in an environment where the actions are already being performed by the slower agent. To achieve this, we consider an action combination function $C:\mathcal{A} \times \mathcal{A} \rightarrow \mathbb{R}$ that outputs the final action that is executed in the environment. We define $C$ as:

\begin{equation}
    C(a^s, a^{f}) := clip(a^s + a^{f}, a_{min}, a_{max})
\end{equation}

where $a_{min}$ and $a_{max}$ respectively denote the minimum and maximum action for the environment, $a^s$ denotes the slow action computed by the pre-trained slow policy, and $a^{f}$ denotes the fast action computed by the fast policy:

\begin{equation}
    \pi^{f}(a^f | s, a^s) := \Pr(A^f_t = a^f | A^s_t = a^s, S_{t}=s)
\end{equation}

Combining actions in this way permits the fast network to act only when needed (since for $a^f = 0$, the action becomes $a^s$), which in turn allows for repetitions of actions by the slow network. This setting provides better exploration for the fast network, as the default slow network behaviour promotes exploring towards reward states, allowing the network to train faster. Additionally, this setting also allows us to easily identify ``emergency states'' where a faster action is required. During evaluation, we threshold the fast action as follows:

\begin{align}
a^f = 
\begin{cases}
    a^f     &\text{if $|C(a^s, a^{f}) - a^s| \geq thresh$} \\
    0       &\text{otherwise} 
\end{cases}
\end{align}

By thresholding the fast action when its influence on the final action is low, we reduce jerky behavior, promoting long smooth actions. If the action has more than one dimension, we set the fast action to the zero vector only if the above condition is true for all dimensions. However, this condition can be adjusted so that it is applied separately for each action dimension. This would be especially useful for settings where each dimension represents a different joint, so that action repetition for each joint is independent.

Finally, we introduce two components to incentivize the fast network to only act when needed and to improve learning:

\begin{enumerate}
    \item \textbf{Fast action penalty (P)}: During pre-training of the slow policy, the fast actions are set to zero. However, after training, if the fast action is set to a non-zero value, the output of the slow network might change drastically since it might be introduced to states that it might not have trained on (as the combined action could produce previously unseen state transitions for the slow policy). If the slow policy is not well generalized, this would hinder the training of the fast policy as it now has to negate drastic changes in the slow policy while trying to maximize return. We find that penalizing any action taken by the fast policy stabilizes training. We implement this by providing a negative reward proportional to the magnitude of the fast policy output. For actions with multiple dimensions, we use the average magnitude over all dimensions.
    
    \item \textbf{Fast state augmentation (PA)}: Another hurdle in training the fast policy is that the effect of its output action is also dependent on the output of the slow policy. Often the combination function would clip the sum of actions to the maximum or minimum possible action, this reduces the sensitivity of the final action to the output of the fast policy. Additionally, the fast policy would have to learn to predict the output of the slow policy to perform well. Thus, to mitigate this, we augment the fast policy inputs with the output of the slow policy. 
    
\end{enumerate}

\subsection{Open-loop temporally layered architecture}
In this setting, we allow partially open loop control by allowing the slow controller to gate the computation of the fast controller. This is achieved by training a slow agent over a pre-trained fast controller. The actions of the slow agent are augmented to include a binary gate output $g \in \{0, 1\}$. Therefore, the slow policy becomes:
\begin{equation}
    \pi^{s}(a^s, g | s) := \Pr(A^s_t = (a^s, g) | S_{t}=s)
\end{equation}

At each step, the final action performed in the environment is:

\begin{equation}
    a_t = g_t \cdot a^s_t + (1-g_t) \cdot a^f_t 
\end{equation}
where $a^f_t$ is the action picked by the pre-trained fast controller at time $t$. Thus, the fast controller computation can be gated if $g=1$. Since the pre-trained fast policy would outperform any slow policy, a naive approach would result in the gate always being set to zero, thus always reverting to the fast policy. To prevent this, we again introduce a reward penalty for picking fast actions. The reward for the slow controller thus becomes:
\begin{align}
r_t = 
\begin{cases}
g_t \cdot r_t + (1-g_t) \cdot r_t \cdot 2        & \text{if $r_t \leq 0$} \\
g_t \cdot r_t + (1-g_t) \cdot r_t \cdot 0.5      & \text{otherwise} 
\label{eq:reward_penalty}
\end{cases}
\end{align}

\begin{table*}[t]
\centering
\resizebox{.95\columnwidth}{!}{
\begin{tabular}{|l|l|l|l|l|l|l|l|l|}
    \hline
    Environment & \multicolumn{4}{|c|}{AUC}  & \multicolumn{4}{|c|}{Avg. Return} \\
    
    \hline \hline
    & TD3 & TempoRL & TLA-O & TLA-C & TD3 & TempoRL & TLA-O & TLA-C\\
    \hline \hline
   Pendulum & 0.86 & 0.858 & \textbf{0.9125}  & \textbf{0.9784} & -150.93 (27.28) & -151.55(26.97) & -153.4(27.67)  & -152.58 (28.22) \\
   \hline
   MountainCarContinuous & 0.5 & \textbf{0.655} & \textbf{0.562} & 0.968 & -0.001 (0.001) & \textbf{56.39(46.05)} & \textbf{37.4(45.81)} & \textbf{94.1 (0.51)}
 \\
   \hline
   InvertedPendulum & 0.973 & 0.952 & 0.526 & \textbf{0.991} & 1000 (0) & 1000(0) & 1000(0)  & 1000 (0) \\
   \hline
   
\end{tabular}
}
\caption{Average normalized AUC and average return results. The standard deviation is reported in the brackets. TLA-O and TLA-C represent the open and closed loop forms of TLA respectively. We highlight all results that are better than the baseline (TD3)}
\label{table1}
\end{table*}

\begin{figure*}[t]
\centering
\includegraphics[width=0.9\columnwidth]{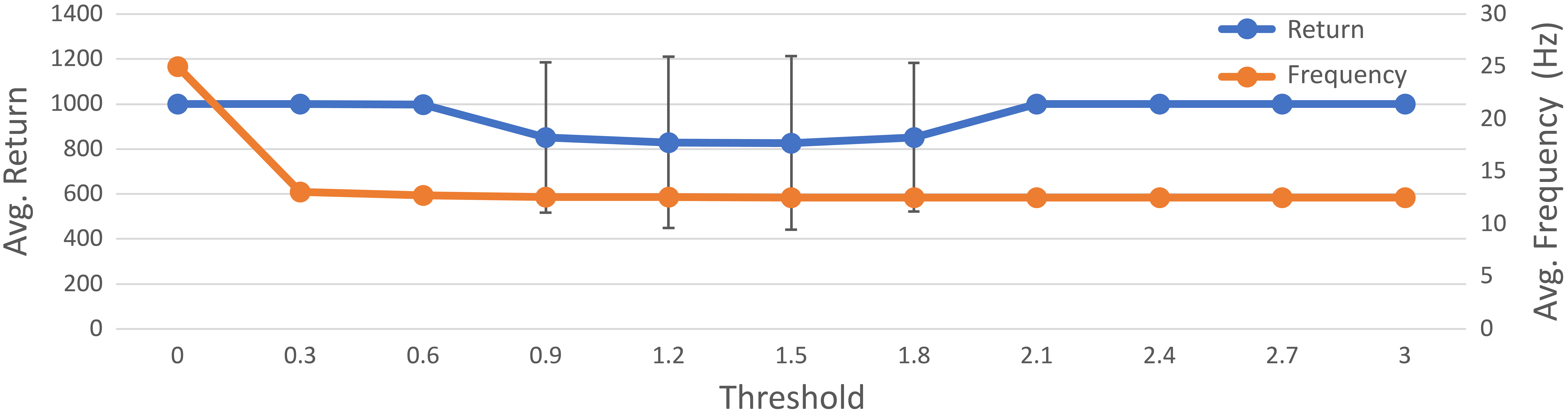} 
\caption{The average return and average action frequency w.r.t. the threshold for TLA-C algorithm in the InvertedPendulum-v2 environment. Increasing the threshold significantly reduces the average actions picked while achieving the same performance. }
\label{fig2}
\end{figure*}

\section{Experiments}

To evaluate our algorithms we measure their performance on six  continuous control tasks including four MuJoCo  tasks\cite{todorov2012mujoco} using the OpenAI Gym environment \cite{openai}. The neural networks are implement and trained using the PyTorch framework \cite{NEURIPS2019_9015}.

\subsection{Turn-based Environments}
Open-loop control is a hard problem. Indeed previous works have generally focused on discrete environments that benefit more from larger step sizes \cite{Biedenkapp2021TempoRLLW, Sharma2017LearningTR}. \citet{Biedenkapp2021TempoRLLW}, who proposed TempoRL, did demonstrate performance on the Pendulum continuous environment, however, they used it as an adversarial example where their algorithm could not outperform the the baseline. Thus, we compare our algorithms to the to plain TD3 and TempoRL on three continuous control environments: Pendulum-v1 and MountainCarContinuous-v0 from the classic control problems and InvertedPendulum-v2 from the MuJoCo tasks of the OpenAI Gym.

For our implementations, we picked our hyperparameters to be consistent with the TD3 implementation \cite{Fujimoto2018AddressingFA}. Both the actor and the critic networks are trained with a learning rate of 3e-4 using the Adam optimizer \cite{Kingma2015AdamAM}. We use two-layer neural networks with 400 and 300 hidden nodes respectively for both the actor and critic.

A noise $\epsilon \sim \mathcal{N}(0, 0.2)$ clipped to (-0.5, 0.5) is added to the actions chosen by the actor for policy smoothing. The actor is updated every 2 iterations. A random policy is implemented for the first 1000 steps for exploration and after that, a noise of $\mathcal{N}(0, 0.1)$ is added  to each action. 

The original TempoRL implementation used the DDPG algorithm \cite{Lillicrap2016ContinuousCW}. For our experiments, we created a TD3 version for comparison.


In order to measure training convergence speed, we measure the normalized average area under the training curves of each task. Additionally we also report the performance on each task. We ran 15 trials for each algorithm on Pendulum-v1 and MountainCarContinuous-v0 tasks and 10 trials for the InvertedPendulum-v2 task. For each task, the TempoRL maximum skip parameter is set to four so that it can at most perform the action for four timesteps. To make TLA comparable, we set the timestep of the fast controller equal to the default timestep and the timestep of the slow controller equal to four times the default timestep. Thus, it can repeat an action for at most four timesteps if the fast controller is gated or its action is thresholded.

The learning curves for each task is shown in Figure \ref{fig:lc}. The results are summarized in Table \ref{table1}. We find that the close-loop-form of TLA (TLA-C) significantly outperforms all other methods. This is because the pre-trained slow controller allows better exploration and more stable learning for the fast controller. The MountainCarContinuous-v0 environment is a difficult task as it requires a lot of exploration without which its policy converges to no action. Here too TLA-C is able to learn the optimal policy.

Surprisingly, the open-loop-form of TLA (TLA-O), converges faster than the TD3 on the Pendulum. We hypothesize that this is due to the larger-timesteps of the TLA-O slow controller that reduce its task horizon. MountainCarContinuous-v0 is difficult for both TempoRL and TLA-O, however, they outperform TD3 due to their ability to take larger steps. Note that the pre-trained fast policy for TLA-O is the same policy as TD3 and thus for MountainCarContinuous-v0, it is trained on top of a sub-optimal policy that always outputs a zero action. Finally, for InvertedPendulum-v2, we again find that TempoRL fails to to outperform TD3. TLA-O's performance on InvertedPendulum-v2 reveals that the learning is unstable and often diverges for the environment. We believe that this is due to the reward always being positive as opposed to the other environments where it is negative for most of the states. The reward penalty, presented in Eq. \ref{eq:reward_penalty}, should be modified for stable learning. We leave that to future work. 

\subsubsection{Action Repetitions and Decisions}
We measure the percentage of actions that are repeated for each method. The action repetition percentage demonstrates how good the method is at "mining" sequences of states for which the agent can perform a long action without reduction in performance. Additionally, the action repetition percentage is also a good measure for the ``jerkiness'' of the policy.

We also measure the average number of decisions for each policy as this is different for different policies. We define a decision as the number of timesteps where a forward pass of any part of the agent is required. Thus, for TLA-C and TD3, the number of decisions is equal to the number of timesteps. For the open-loop algorithm, the number of decisions is decreased resulting in compute savings. However, it should be noted that the number of decisions does not provide complete information of the compute savings for TLA-O since each decision may make use of either the slow controller, fast controller or both. This adaptive compute capability is not present in TempoRL where for each decision two networks are required: one for the action and one for the number of timesteps.

\begin{table*}[t]
\centering
\begin{tabular}{|l|l|l|l|l|l|l|l|l|}
    \hline
    Environment & \multicolumn{4}{|c|}{Action Repetitions}  & \multicolumn{4}{|c|}{Decisions} \\
    
    \hline \hline
    & TD3 & TempoRL & TLA-O & TLA-C & TD3 & TempoRL & TLA-O & TLA-C\\
    \hline \hline
   Pendulum & $6.04\%$ & 27.37$\%$ & 14.70\% & 70.96\% & 200.00 & 151.67  &179.84 &200.00
 \\
  
   \hline
   InvertedPendulum & 1.12\% & 55.29\% & 58.86\% & 74.9\% & 1000 & 445.2 & 802.56  & 1000 \\
   \hline
   
\end{tabular}
\caption{Average action repetition percentage and decisions.}
\label{table:actions}
\end{table*}

\subsubsection{Threshold vs. Return}

The TLA architecture has two different controllers trained at two different response times. In the TLA-C architecture, we threshold the fast action in order to promote action repetition. Thus by gating the actions, we are effectively reducing the response time of TLA. When the threshold for the fast action is changed, it also changes the average action frequency of TLA.  Fig. \ref{fig2} demonstrates the variance of the average return and average action frequency with respect to the threshold for the InvertedPendulum-v2 environment. Similar plots for other MuJoCo environments are added in the appendix. However, in this particular case, the pre-trained slow policy also achieves the maximum possible reward. The slow policy is trained at 12.5Hz and the fast policy is trained at 25Hz. Interestingly, TLA performs optimally for all frequencies except in the frequency range (12.5, 12.55]. This might indicate the temporal generalization gap of TLA, as it performs well when near the frequencies it is trained on but the performance deteriorates in the middle.

\subsubsection{Temporal context of states}
\begin{figure}[t]

\includegraphics[width=0.1\textwidth]{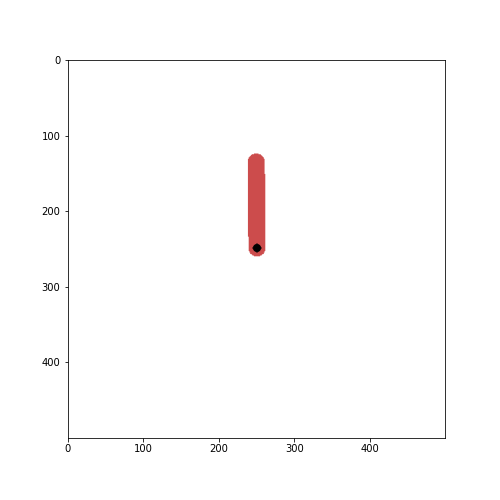}
\includegraphics[width=0.1\textwidth]{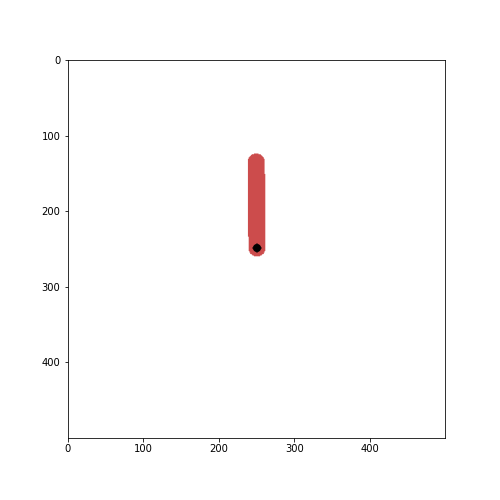}
\includegraphics[width=0.1\textwidth]{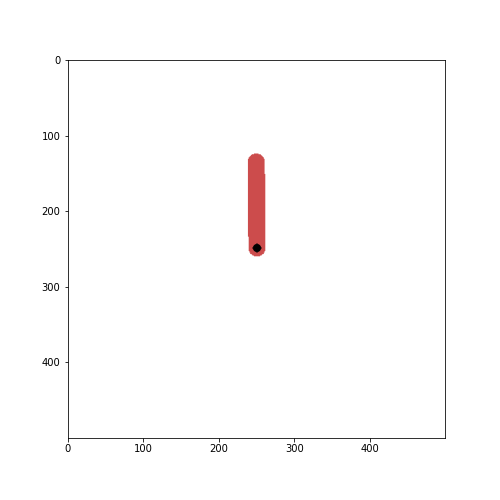}
\includegraphics[width=0.1\textwidth]{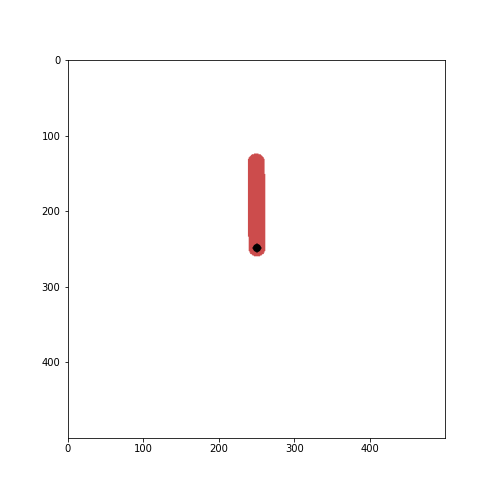}
\includegraphics[width=0.1\textwidth]{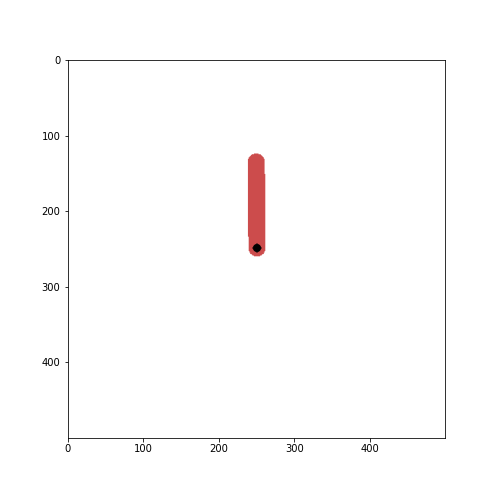}
\includegraphics[width=0.1\textwidth]{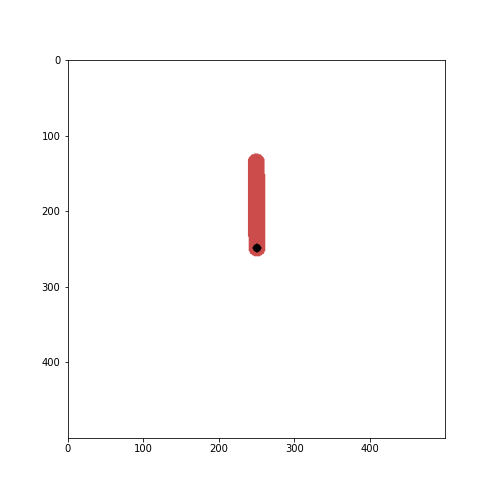}
\includegraphics[width=0.1\textwidth]{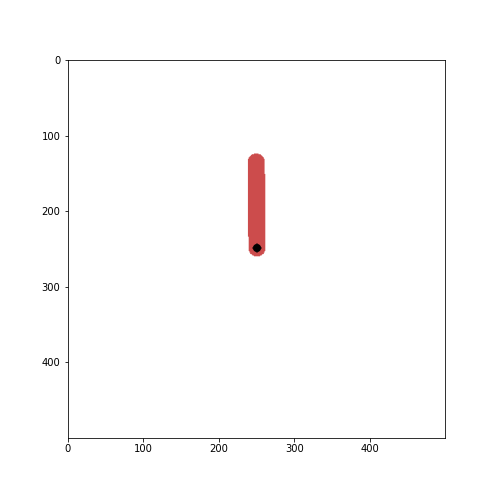}
\includegraphics[width=0.1\textwidth]{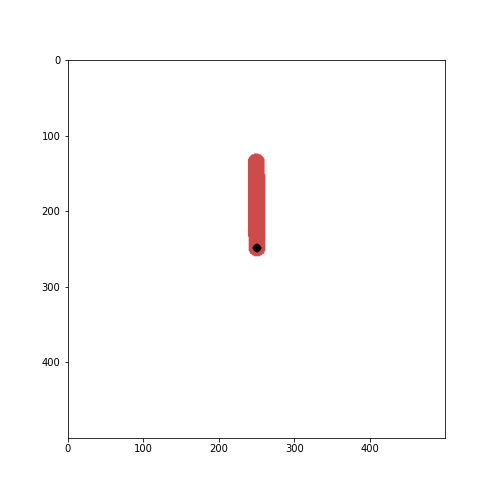}
\caption{Examples of states in Pendulum-v1 environment where the fast controller is activated for the TLA-C setting.}
\label{states}
\end{figure}

The TLA-C method can be used to automatically sample temporally important states where a fast action is required. Fig \ref{states} demonstrates some such sampled states from the Pendulum-v1 environment. We find that the fast network is only activated when the pendulum is upright as that is when fine-grained control is critical to maximize rewards. 

\subsection{Real-time Environments}
Since the temporally layered architecture is distributed. It can be applied in a setting where each layer is on a different system. This is similar to biological reflexes that are independent of the brain but jointly define the control policy with the brain. Each layer focuses on a different temporal context and has a different reaction time. This allows for a fast action, that is seldom activated, layered on top of a slower action that is outputted by a controller with slower response time. 

We demonstrate distributed control in simulated real-time environments, implementing the real-time setting from \citet{Chen2021DelayAwareMR}. The simulated delay is equal to the timestep of each layer. The states of the environments are augmented by the previous action picked by the agent, which are performed in the environment after a delay of one timestep. For TLA, the actions picked by the slow controller will be delayed longer since it has a larger timestep. 

For a fair comparison, the fast time-step is set equal to the original time-step defined in the environment. The slow layer time-step is set to double the length of the fast time-step. Thus, the TLA agent can only act as fast as the benchmark. During evaluation, the threshold for gating the fast action was set to 0.3 $\times$ max action. However, tuning the threshold for each individual environment would result in a better performance or efficiency. 

Our results are presented in Table \ref{table3}. TLA has a higher or equal average return than TD3 on all tested environments. Additionally, TLA also has significantly higher action repetition in all the environments. Action repetition in complex continuous environment is very rare. But using the closed-loop variant of TLA, we are able to increase action repetition while maintaining the performance. Finally, TLA is able to learn faster in two out of the three environments and has a higher normalized AUC. We include learning curves for the experiments in the supplementary material.

\begin{table*}[t]
\centering
\resizebox{.95\columnwidth}{!}{
\begin{tabular}{|l|p{2.2cm}|l|l|p{1.3cm}|l|l|}
    \hline
    Environment & \multicolumn{3}{|c|}{TLA}  & \multicolumn{3}{|c|}{TD3} \\
    
    \hline \hline
    & AUC & Return & AR & AUC & Return & AR\\
    \hline
   InvertedDoublePendulum-v2 & \textbf{0.979} & \textbf{9358.94 $\pm$ 0.82} & \textbf{49.41\%} & 0.966 & 9358.48 $\pm$ 2.56 & 1.05\%\\
  \hline
   Hopper-v2 & \textbf{0.657} & \textbf{3443.21 $\pm$ 131.61} & \textbf{5.57\%} & 0.413 & 3032.25 $\pm$ 262.81 & 0\%\\
   \hline
   Walker2d-v2  & 0.665 & \textbf{3694.04 $\pm$ 128.58} & \textbf{6.83\%} & \textbf{0.678} & 3233.77 $\pm$ 895.33 & 0.01\%\\
   \hline
   
\end{tabular}
}
\caption{Average return over 5 trials and the corresponding AUC and action repetition percentage (AR) and response times. The best return, AR and AUC are in bold. $\pm$ corresponds to a single standard deviation over trials.}
\label{table3}
\end{table*}

\section{Discussion and Future work}

For our TLA design, we drew upon the brain's ability to utilize environmental information and respond at multiple timescales. For instance, the brain is able to adjust reaction time, strategy, and future planning based on the environment's energy constraints and task demands. As well as balance reflexes against planned motor movements to achieve accurate and efficient responses in a changing environment. We demonstrate that the TLA's design enables it to detect emergency situations quickly and survive in unknown environments for long periods of time. 

The brain's design allows it to process environmental information in parallel while using distributed control to make choices. This design enables the brain to accurately and efficiently adapt to a changing environment (for a review of human decision-making systems, please see: {\cite{van2012information}}). The ability to process sensory information while simultaneously picking actions has enabled many species to navigate and evolve in a large array of complex environments. Thus, to create adaptable and intelligent design, researchers should look to biological designs and utilize these neural architectures to build truly intelligent, adaptive systems.

The Affordance Competition Hypothesis \cite{cisek2007cortical}, a specific theory of motor planning, was leveraged for TLA but can be expanded on in future designs to continue to improve TLA and inspire new architectures.  This theory of motor-planning provides a framework to explain how the mammalian brain creates a motor plan based on a changing environment. This theory predicts that the animal has a representation of multiple motor plans and chooses the one that out-competes the other plans. These  action plans can have multiple timescales, goals, action costs and environmental information included in their representations. This model predicts that the animal chooses the most appropriate action based on the environment's demands and the animal's goal. This theory also provides a clear and behavior-supported model for how discrete goal-directed movements are incorporated into a fluid movement plan as the environment moves forward in time (e.g., if a lion will continue to hunt a gazelle that suddenly changed direction or if the lion will give up and wait for a new prey). In this theory, the brain creates multiple motor plans that compete to be executed and one wins through a comparison competition to the other motor plans' cost/benefit ratio.

 Another example of the brain's ability to adapt to environmental demands can be see under reduced gravity where the human gait switches from a walk to a run at slower speeds \cite{kram1997effect} On the other hand, as shown in this work, the same environment might be perceived differently under different response times. In nature, this phenomenon results in a reduction in the relative locomotion speed as the response times get slower and the body sizes get larger \cite{IriarteDaz2002DifferentialSO}. A truly intelligent AI must be able to navigate many different environments that might require different response frequencies. We believe that our work is an important step towards a temporally intelligent AI.
 
 The slower layers of TLA benefit from a shorter task horizon and better exploration since they can be considered macro-actions. Each additional faster layer can be intuitively interpreted as adding temporal details to the policy. Since the faster layers only activate when there is a higher expected return than not acting, TLA can potentially be used to better understand complex environments: identifying states that require a faster response time. TLA automatically categorizes the states that require urgent action and states that do not. In biology, evolution has naturally given rise to such a temporally layered control with reflexes that activate in urgent situations. Like reflexes, that are passed down to the next generation, the faster layers of TLA can be used for knowledge distillation. Yet, similar to our TLA-O approach that is trained from fast to slow, reflexes are faster pre-trained layers and slower layers are trained on top of them. However, like macro-actions, some reflexes also aid learning due to reduced exploration. We leave a more in-depth exploration of this to future work.

Our gating of the fast controller in TLA-O setting is similar to techniques used in early-exit architectures \cite{scardapane2020should}. In contrast to the early-exit scenario, in a real-time control setting, the output of the network needs to be different at different times. Thus, the agent needs to be aware of its own processing and actuation delay when picking an output. Additionally when ``answering early'' the action needs to be changed accordingly. We demonstrated this in our real-time experiments and hope to pave the way for future research in temporally intelligent RL agents that can ``understand'' the impact of time on optimal action.

\section{Conclusion}
In this work, we presented Temporally Layered Architecture (TLA), a framework for distributed, adaptive response time in reinforcement learning. The framework allows the RL agent to achieve smooth control in a real-time setting using a slow controller while a fast controller monitors and intervenes as required. Additionally, we demonstrated an alternative setting where the slow controller can gate the fast controller, activating it only when required for efficient control. We demonstrate faster convergence and more action repetition in the closed-loop approach and fewer decision and faster convergence in the partially-open loop approach. Additionally, we demonstrate in a real time setting, where processing and actuation delays are taken into account, and show that our approach outperforms the current approaches in the delayed setting while picking fewer actions. Our work demonstrates that a temporally adaptive approach has similar benefits for AI as has been demonstrated in biology and is an important direction for future research in artificially intelligent control. 



\section*{Acknowledgments}
This material is based upon work partially supported by the
Defense Advanced Research Projects Agency (DARPA) under Agreement No. HR00112190041.  
The information contained in this work does not
necessarily reflect the position or the policy of the Government.



\bibliography{references}


\end{document}